\documentclass{bmvc2k}

\usepackage{booktabs}
\usepackage{wrapfig}
\usepackage{multirow}
\usepackage{amssymb}
\usepackage{amsmath,amsfonts,bm}

\def\eqref#1{equation~\ref{#1}}
\def\1{\bm{1}}

\def\vf{{\bm{f}}}

\def\vh{{\bm{h}}}

\def\vp{{\bm{p}}}
\def\vq{{\bm{q}}}

\def\vs{{\bm{s}}}

\def\vx{{\bm{x}}}
\def\vy{{\bm{y}}}

\DeclareMathAlphabet{\mathsfit}{\encodingdefault}{\sfdefault}{m}{sl}
\SetMathAlphabet{\mathsfit}{bold}{\encodingdefault}{\sfdefault}{bx}{n}

\DeclareMathOperator*{\argmax}{arg\,max}

\usepackage{subcaption}
\usepackage{amsthm}
\usepackage{enumitem}
\usepackage{xcolor, soul}
\usepackage{todonotes}
\usepackage{fancyhdr,graphicx,amsmath,amssymb}
\usepackage[ruled,vlined,linesnumbered]{algorithm2e}
\usepackage{cleveref}

\title{SSR: An Efficient and Robust Framework for Learning with Unknown Label Noise}

\addauthor{Chen Feng}{https://sites.google.com/view/mr-chenfeng}{1}
\addauthor{Georgios Tzimiropoulos}{https://ytzimiro.github.io/}{1}
\addauthor{Ioannis Patras}{https://sites.google.com/view/ioannispatras}{1}

\addinstitution{
 School of Electronic Engineering and Computer Science\\
 Queen Mary University of London\\
 London, UK
}

\runninghead{Chen et al}{SSR: An Efficient and Robust Framework for LULN}

\begin{document}

\maketitle
\begin{abstract}
Despite the large progress in supervised learning with neural networks, there are significant challenges in obtaining high-quality, large-scale and accurately labelled datasets. In such a context, how to learn in the presence of noisy labels has received more and more attention. As a relatively complex problem, in order to achieve good results, current approaches often integrate components from several fields, such as supervised learning, semi-supervised learning, transfer learning and resulting in complicated methods. Furthermore, they often make multiple assumptions about the type of noise of the data. This affects the model robustness and limits its performance under different noise conditions. 
In this paper, we consider a novel problem setting, \textit{\textbf{L}earning with \textbf{U}nknown \textbf{L}abel \textbf{N}oise}~(\textbf{LULN}), that is, learning when both the degree and the type of noise are unknown. Under this setting, unlike previous methods that often introduce multiple assumptions and lead to complex solutions, we propose a simple, efficient and robust framework named \textit{\textbf{S}ample \textbf{S}election and \textbf{R}elabelling}~(\textbf{SSR}), that with a minimal number of hyperparameters achieves SOTA results in various conditions. At the heart of our method is a sample selection and relabelling mechanism based on a \textbf{n}on-\textbf{p}arametric \textbf{K}NN classifier~(NPK) $g_q$ and a \textbf{p}arametric \textbf{m}odel \textbf{c}lassifier~(PMC) $g_p$, respectively, to select the clean samples and gradually relabel the noisy samples.
Without bells and whistles, such as model co-training, self-supervised pre-training and semi-supervised learning, and with robustness concerning the settings of its few hyper-parameters, our method significantly surpasses previous methods on both CIFAR10/CIFAR100 with synthetic noise and real-world noisy datasets such as WebVision, Clothing1M and ANIMAL-10N. Code is available at \url{https://github.com/MrChenFeng/SSR_BMVC2022}.
\end{abstract}

\section{Introduction}
It is now commonly accepted that supervised learning with deep neural networks can provide excellent solutions for a wide range of problems, so long as there is sufficient availability of labelled training data and computational resources. However, these results have been mostly obtained using well-curated datasets in which the labels are of high quality. In the real world, it is often costly to obtain high-quality labels, especially for large-scale datasets. A common approach is to use semi-automatic methods to obtain the labels (e.g. ``webly-labelled'' images where the images and labels are obtained by web-crawling). While such methods can greatly reduce the time and cost of manual labelling, they also lead to low-quality noisy labels. 

\begin{wrapfigure}{r}{0.4\textwidth}
%\vspace{-5mm}
  \begin{center}
    \includegraphics[width=0.4\textwidth]{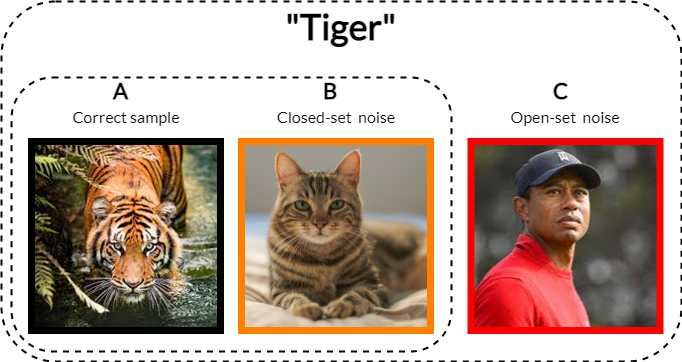}
  \end{center}
    \caption{Different ``tigers''.}
        % \caption{Different ``tigers'' in an animal dataset.}
  \label{fig:noise_example}
  %\vspace{-5mm}
\end{wrapfigure}

% two main types of methods
%Formally speaking, in supervised classification problems, noisy samples usually come from two main categories: closed-set noise where the true labels belong to one of the given classes (Set B in~\cref{fig:noise_example}) and open-set noise where the true labels do not belong to the set of labels of the classification problem (Set C in~\cref{fig:noise_example}). To deal with different noises, two main types of methods have been proposed, which we define here as probability-consistent methods and probability-approximate methods.
In such settings, noise is one of the following two types: closed-set noise where the true labels belong to one of the given classes (Set B in~\cref{fig:noise_example}) and open-set noise where the true labels do not belong to the set of labels of the classification problem (Set C in~\cref{fig:noise_example}). To deal with different types of noise, two main types of methods have been proposed, which we name here as probability-consistent methods and probability-approximate methods.
Probability-consistent methods usually model noise patterns directly and propose corresponding probabilistic adjustment techniques, e.g., robust loss functions~\citep{mae, generalized_cross_entropy, symmetric_cross_entropy} and noise corrections based on noise transition matrix~\citep{noiseadaptation}. However, accurate modelling of noise patterns is non-trivial, and often cannot even model open-set noise. Also, due to the necessary simplifications of probabilistic modelling, such methods often perform poorly with heavy and complex noise. 
More recently, probability-approximate methods, that is methods that do not model the noise patterns explicitly become perhaps the dominant paradigm, especially ones that are based on sample selection. Earlier methods often reduce the influence of noise samples by selecting a clean subset and training only with it~\citep{coteaching, coteaching+, mentornet, whentohow}. Recent methods tend to further employ semi-supervised learning methods, such as MixMatch~\citep{mixmatch}, to fully explore the entire dataset by treating the selected clean subset as labelled samples and the non-selected subset as unlabeled samples~\citep{dividemix, moit}. These methods, generally, do not consider the presence of open-set noise in the dataset, since most current semi-supervised learning methods can not deal with open-set noise appropriately. To address this, several methods~\citep{evidentialmix, ngc} extend the sample selection idea by further identifying the open-set noise and excluding it from the semi-supervised training. 

In general, the above methods make assumptions about the pattern of the noise, such as the confidence penalty specifically for asymmetric noise in DivideMix~\citep{dividemix}. However, these mechanisms are often detrimental when the noise pattern does not meet the assumptions -- for example, explicitly filtering open-set noise in the absence of open-set noise may result in clean hard samples being removed. Furthermore, due to the complexity of combining multiple modules, the above methods usually need to adjust complex hyperparameters according to the type and degree of noise. 
% To cope with more complicated noise, regularizations and modifications are proposed for better performance, which, however, seriously affects the robustness and simplicity of the model. 

%In general, the above methods make assumptions about the pattern of the noise, such as the confidence penalty specifically for asymmetric noise in DivideMix~\citep{dividemix}. However, these mechanisms are often detrimental when the noise pattern does not meet the assumptions -- for example, filtering open-set noise in the absence of open-set noise may result in clean hard samples being removed. At the same time, due to the complexity of combining multiple modules, the above methods usually need to adjust complex hyperparameters according to the type and degree of noise. To cope with more complicated noise, increasingly regularizations and modifications are proposed for better performance, which, however, seriously affects the robustness and simplicity of the model.

In this paper, we consider a novel problem setting --- \textit{\textbf{L}earning with \textbf{U}nknown \textbf{L}abel \textbf{N}oise}~(\textbf{LULN}), that is, learning when both the degree and the type of noise are unknown.
Striving for simplicity and robustness, we propose a simple method for \textbf{LUNL}, namely \textit{\textbf{S}ample \textbf{S}election and \textbf{R}elabelling}~(\textbf{SSR})~(\cref{3_3}), with two components that are clearly decoupled: a selection mechanism that identifies clean samples with correct labels, and a relabelling mechanism that aims to recover correct labels of wrongly labelled noisy samples. These two major components are based on the two simple and necessary assumptions for \textbf{LULN}, namely, that samples with highly-consistent annotations with their neighbours are often clean, and that very confident model predictions are often trustworthy. Once a well-labelled subset is constructed this way we use the most basic supervised training scheme with a cross-entropy loss. Optionally, a feature consistency loss can be used for all data so as to deal better with open-set noise. 

Without bells and whistles, such as semi-supervised learning, self-supervised model pre-training and model co-training, our method is shown to be robust to the values of its very few hyperparameters through extensive experiments and ablation studies and to consistently outperform the state-of-the-art by a large margin in various datasets.
%Our work serves as a simple yet highly effective baseline, while also inspiring people to rethink the practical effects of various modification techniques and regularizations currently proposed.
%We then use the selected clean subset and cross-entropy loss for the basic supervised training. Optionally, we propose a feature consistency loss to further utilise even open-set noise. Without bells and whistles, such as semi-supervised learning, model pre-training and model cotraining, our method proves to be robust to the values of its very few hyperparameters through extensive experiments and ablation studies and consistently outperforms state-of-the-art by a large margin in various datasets. Our work serves as a simple yet highly effective baseline, while also inspiring people to rethink the practical effects of various modification techniques and regularizations currently proposed.

\section{Related Works}
This paper mainly focuses on the probability-approximate methods, especially methods based on sample selection. For a detailed introduction to probability-consistent methods described above, please refer to the review papers~\citep{review1, review2}. We note that we do not consider utilizing an extra clean validation dataset, such as meta-learning-based methods~\citep{mwnet, famus} do. 
%\vspace{-5pt} 

\paragraph{Clean sample selection} Most sample selection methods fall into two main categories: 
\begin{itemize}[itemsep=0em]
\item \textit{Prediction-based methods.} Most of the recent sample selection methods do so, by relying on the predictions of the model classifier, for example on the per-sample loss~\citep{lossmodellingbmm, dividemix} or model prediction~\citep{selfie, whentohow}. However, the prediction-based selection is often unstable and easily leads to confirmation bias, especially in heavy noise scenarios. A few works focus on improving the sample selection quality of these methods~\citep{xia2021sample, zhou2020robust}. To identify open-set noise, several methods utilize the Shannon entropy of the model predictions of different samples~\citep{evidentialmix, albert2022addressing}. Open-set noise samples that do not belong to any class should have a relatively average model prediction (larger entropy value).

\item \textit{Feature-based methods.} Instead of selecting samples based on the model prediction, some works try to utilize the feature representations for sample selection. \citet{topofilter,ngc} try to build a KNN graph and identify clean samples through connected sub-graphs. \citet{deep-knn} selects clean samples with a KNN classifier in the prediction logit space, while \citet{moit} proposes an iterative KNN to alleviate the effect of noisy labels.
\end{itemize}

Our work falls in the second category. However, unlike existing methods that use complex variants of neighbouring algorithms, in our pursuit of simplicity and robustness, we use the simplest KNN classification and show that this is sufficient. % and subsequently a simple training scheme.

\paragraph{Fully exploiting the whole dataset} To fully utilise the whole dataset during training and more specifically the non-selected subset, recent methods usually apply semi-supervised training methods (e.g., MixMatch~\citep{mixmatch}), by considering the selected subset as labelled and the non-selected subset as unlabeled~\citep{dividemix}. However, most current semi-supervised learning methods can not deal with open-set samples properly. How to properly do semi-supervised learning in this setting is often referred to as open-set semi-supervised learning~\citep{MTC, openmatch}. In this paper, instead of adopting complex semi-supervised learning schemes, we adopt a simple relabeling and selection scheme in order to construct a clean and well-labelled subset and then train with a simple cross-entropy loss on the clean, well-labelled set and optionally, with a feature consistency loss on the whole dataset that possibly contains open-set noise and samples that cannot be well relabelled.

\section{Methodology}
\label{3}
\subsection{Problem formulation}
\label{3_1}
Let us denote with $\mathcal{X} = \{\bm{x}_i\}_{i=1}^N, \bm{x}_i \in R^d$, a training set with the corresponding one-hot vector labels $\mathcal{Y} = \{\bm{y}_i\}_{i=1}^N, \vy_i \in \{0, 1\}^M$, where $M$ is the number of classes and $N$ is the number of samples.
For convenience, let us also denote the label of each sample $\vx_i$ corresponding to the one-hot label vector $\vy_i$ as $l_i = \arg_j[y_i(j)=1] \in \{1, ..., M\}$. 
%Finally, let us denote the true labels with $\mathcal{Y}' = \{\vy'_i\}_{i=1}^N$. Clearly, for an open-set noisy label it is the case that $\vy'_i \neq \vy_i, \vy'_i \notin \{0, 1\}^M$, while for closed-set noisy samples $\vy'_i \neq \vy_i, \vy'_i \in \{0, 1\}^M$. 
Finally, let us denote the true labels with $\mathcal{Y}' = \{\vy'_i\}_{i=1}^N$. Clearly, for an open-set noisy label it is the case that $\vy'_i \neq \vy_i, \vy'_i \notin \{0, 1\}^M$, while for closed-set noisy samples $\vy'_i \neq \vy_i, \vy'_i \in \{0, 1\}^M$.

% Aiming to learn with the noise-agnostic dataset, 
We view the classification network as an encoder $f$ that extracts a feature representation and a \textbf{p}arametric \textbf{m}odel \textbf{c}lassifier~(PMC) $g_p$ that deals with the classification problem in question. We also define a \textbf{n}on-\textbf{p}arametric \textbf{K}NN classifier~(NPK) $g_q$ based on the feature representations from encoder $f$.
For brevity, we define $\vf_i \triangleq f(\vx_i)$ as the feature representation of sample $\vx_i$, and $\vp_i \triangleq g_p(\vf_i)$ and $\vq_i \triangleq g_q(\vf_i)$ as the prediction vectors from PMC $g_p$ and NPK $g_q$, respectively. Following recent works~\citep{dividemix, ngc, moit, coteaching, coteaching+}, we adopt an iterative scheme for our method consisting of two stages: 1.~sample selection~(\cref{alg:sel}) and relabelling~(\cref{alg:re}), and 2.~model training~(\cref{alg:train}) in \textbf{Algorithm 1}. %\cfnote{They will know what happened after reading the method. I don't think we have space to repeat here...}

% Note some methods do not strictly distinguish the two stages but dynamically select clean samples in an online style. We believe that the online method does not have explicit advantages, while the offline method is clearer and simpler, and can be further simplified by reducing the steps of the first stage thus reducing the additional computational cost.

\begin{figure}[t]
    \begin{center}
    \includegraphics[width=1.0\textwidth]{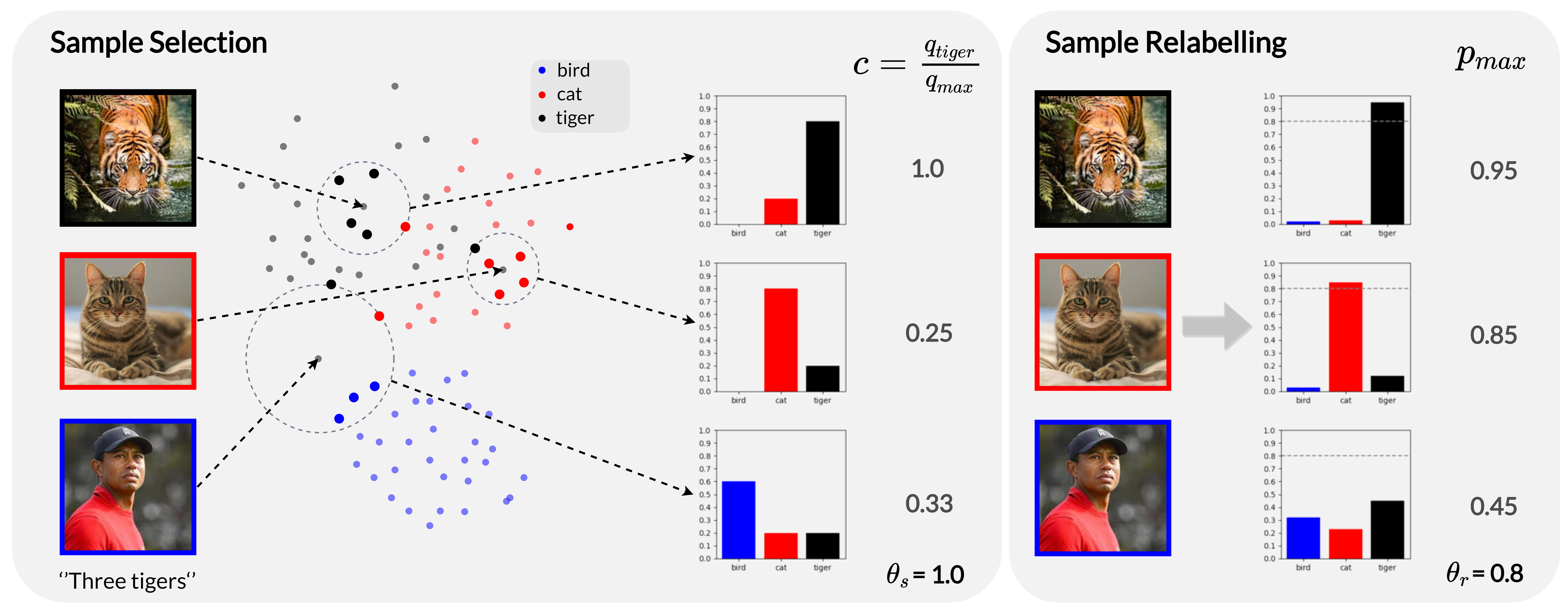}
    \end{center}
    % \caption{A toy example of \textbf{SSR} with a noisy animal dataset.}
    \caption{A toy example of \textbf{SSR}~(\cref{3_3}) with a noisy animal dataset.}
    \label{fig:ssr}
\end{figure}

% \begin{minipage}{20cm}
\begin{algorithm}[H]
\SetAlgoLined
\SetKwInOut{Input}{Input}
\Input{dataset $(\mathcal{X},\ \mathcal{Y})$, sample selection threshold $\theta_s$, sample relabelling threshold $\theta_r$, weight of feature consistency loss $\lambda$, max epochs $T$}
 \While{$i<T$}{
    Generate $(\mathcal{X},\ \mathcal{Y}^r)$ with \cref{eqr} \tcc*[r]{sample relabelling} \label{alg:re}
    Generate $(\mathcal{X}_c,\ \mathcal{Y}^r_c)$ with \cref{eq3} and \cref{eq4} \tcc*[r]{sample selection} \label{alg:sel}
    Model training with \cref{eq10} \tcc*[r]{model training} \label{alg:train}
    % $L = L_{ce}~\cref{eq8}+ \lambda L_{fc}~\cref{eq9}$ \tcc*[r]{model training}
}
\caption{\textbf{SSR}. }
\label{alg}
\end{algorithm}
% \end{minipage}

\subsection{Sample selection and relabelling}
\label{3_3}
For a better exposition, we first introduce our sample selection mechanism. Please note, that we actually relabel the samples before each sample selection. 

\paragraph{Clean sample selection by balanced neighbouring voting}

% \subsubsection{Clean sample selection by balanced neighbouring voting}
\label{3_3_1}
Our sample selection is based on the consistency, as quantified by a measure $c_i$, between the label $\vy_i^r$~\footnote{Please note, we use the labels $\mathcal{Y}^r$~(\cref{eqr}) that a relabelling mechanism provides as mentioned above.} of sample $\vx_i$ and an (adjusted) distribution, $\vq_i$, of the labels in its neighbourhood in the feature space. More specifically, let us denote the similarity between the representations $\vf_i$ and $\vf_j$ of any two samples $\vx_i$ and $\vx_j$ by $\vs_{ij}, i,j=1,...,N$. By default, we used the cosine similarity, that is, $\vs_{ij} \triangleq \frac{\vf_i^T \vf_j}{ \| \vf_i\|_{\small 2} \| \vf_j\|_{\small 2}}$.
Let us also denote by $N_i$ the index set of the $K$ nearest neighbours of sample $\vx_i$ in $\mathcal{X}$ based on the calculated similarity. Then, for each sample $\vx_i$, we can calculate the normalised label distribution $\vq'_i =\frac{1}{K}\sum_{n \in N_i} \vy_n^r$ in its neighbourhood, and a balanced version, $\vq_i \in R^M$, of it that takes into consideration/compensates for the distribution $\bm{\pi} = \sum^N_{i=1} \vy_i^r$ of the labels in the dataset. More specifically,
\begin{equation} \label{eq3}
    \vq_i= \bm{\pi}^{-1} \vq'_i,
\end{equation}
where we denote with $\bm{\pi}^{-1}$ the vector whose entries are the inverses of the entries of the vector $\bm{\pi}$ --- in this way we alleviate the negative impact of possible class imbalances in sample selection.

%We first introduce how to calculate $\vq_i$, or say, the principle of NPK $g_q$ as it is non-trivial. Let us denote the similarity between the representations $\vf_i$ and $\vf_j$ of any two samples $\vx_i$ and $\vx_j$ by $\vs_{ij}, i,j=1,...,N$. By default, we used the cosine similarity, that is, $\vs_{ij} \triangleq \frac{\vf_i^T \vf_j}{ \| \vf_i\|_{\small 2} \| \vf_j\|_{\small 2}}$.
%Let us also denote by $N_i$ the index set of the $K$ nearest neighbours of sample $\vx_i$ in $\mathcal{X}$ based on the calculated similarity. Then, for each sample $\vx_i$, we calculate the balanced label distribution $\vq_i \in R^M$ in its neighbourhood in the feature space, as the normalized sum of its neighbours' labels. More specifically,
%\begin{equation} \label{eq3}
 %   \vq_i= \bm{\pi}^{-1} \vq'_i,
%\end{equation}
%where $\vq'_i =\frac{1}{K}\sum_{n \in N_i} \vy_n^r$~\footnote{Please note, we use the labels $\mathcal{Y^r}$~(\cref{eqr}) that a relabelling mechanism provides as mentioned above.} and $\bm{\pi} = \sum^N_{i=1} \vy_i^r$. With a slight abuse of notation, we denote by $\bm{\pi}^{-1}$ the vector whose entries are the inverses of the entries of the vector $\bm{\pi}$ of the class probabilities in the whole dataset --- in this way we compensate for possible class imbalances. 

The vector $\vq_i$ can be considered as the (soft) prediction of the NPK $g_q$ classifier. We then, define a consistency measure $c_i$ between the sample's label $l_i^r = \argmax_j \vy_i^r(j)$ (\cref{3_1}) and the prediction $\vq_i$ of the NPK as %\footnote{Please, note that $l_i^r = \argmax_j \vy_i^r(j)$ -- cf. section 3.1.}
%\footnote{Please refer to section 3.1 for the correspondence of $l_i^r$ and $\vy_i^r$.}
\begin{equation} \label{eq4}
    c_i = \frac{\vq_i(l_i^r)}{\max_j \vq_i(j)}, % \text{where} l_i^r = \argmax_j \vy_i^r(j)
\end{equation}
that is the ratio of the value of the distribution $\vq_i$ at the label $l_i^r$~(\cref{eqr}) divided by the value of its highest peak $\max_j \vq_i(j)$. 
Roughly speaking, a high consistency measure $c_i$ at a sample $\vx_i$ means that its neighbours agree with its current label $l_i^r$ --- this indicates that $l_i^r$ is likely to be correct. By setting a threshold $\theta_s$ to $c_i$, a clean subset $(\mathcal{X}_c,\ \mathcal{Y}^r_c)$ can be extracted. In our method, we set $\theta_s=1$ by default, that is, we consider a sample $\vx_i$ to be clean only when its neighbours' voting $\vq_i$ is consistent with its current label $\vy_i^r$.

% \subsubsection{Noisy sample relabelling by classifier thresholding}

\paragraph{Noisy sample relabelling by classifier thresholding}
\label{3_3_2}

%The NPK $g_q$ prediction $\vq_i$ of a sample $\vx_i$ constructed by~\cref{eq3} is to some degree affected by the noisy labels of its neighbours. To alleviate this, and also introduce more closed-set samples into training, we propose to utilize PMC $g_p$ for sample relabelling. More specifically, with the PMC $g_p$ prediction $\vp_i$ of sample $\vx_i$ and its original annotated label $l_i$, we then "relabel" all samples by thresholding $\vp_i$ as:
%The NPK $g_q$ prediction $\vq_i$ of a sample $\vx_i$ constructed by~\cref{eq3} is to some degree affected by the noisy labels of its neighbours. To alleviate this, and also introduce more closed-set samples into training, 

Our sample relabelling scheme aims at adding well-labelled samples to the training pool and is based on the PMC classifier $g_p$. Specifically, we "relabel" all samples for which the classifier is confident, that is all samples $i$ for which the prediction $\vp_i$ of the classifier PMC $g_p$ exceeds a threshold $\theta_r$. Formally,
%with PMC $g_p$ prediction $\vp_i$ of sample $\vx_i$, we "relabel" all samples by thresholding $\vp_i$ as:
\begin{equation} \label{eqr}
    l_i^r =\left\{
\begin{aligned}
& \argmax_l \vp_i(l), &~\max_l \vp_i(l) > \theta_r \\
& l_i, &~\max_l \vp_i(l) \leq \theta_r \\
\end{aligned}
\right.
\end{equation}

Please note, that similarly to~\cref{3_1}, we denote the one-hot label corresponding to $l_i^t$ as $\vy_i^t$ --- this will be used in~\cref{eq3}. 
By setting a high $\theta_r$, a highly confident sample $\vx_i$ will be relabelled --- this can in turn further enhance the quality of sample selection. Note, that this scheme typically avoids mis-relabelling open-set noise samples as those tend not to have highly confident predictions. In this way, our method can deal with open-set noise datasets effectively even though we do not explicitly propose a mechanism for them.
% \subsubsection{On choices of sample selection and relabelling}

\paragraph{On choices of sample selection and relabelling}
\label{3_3_3}
In this work, we utilize two different classifiers for the two stages of our scheme: for sample selection the NPK $g_q$ based on nearest neighbours in the feature space  ((\cref{eq3} and \cref{eq4}) and for sample relabelling the PMC $g_p$ classifier (\cref{eqr}). Here, we justify/comment on this choice.
% \paragraph{Why not PMC $g_p$ for sample selection?}

\textit{Why not PMC $g_p$ for sample selection?}
\quad Most previous works rely on the PMC $g_p$ itself to select clean samples, i.e., typically, samples with small losses. However, it is well known that such methods are not robust to complex and heavy noise. Also, these methods often require a warmup stage before sample selection --- this requires prior knowledge about the difficulty and the noise ratio of the dataset. For example, a warmup stage of 50 epochs under heavy noise on the CIFAR10 dataset may result in overfitting before sample selection, while a warmup stage of 5 epochs on the mildly noisy CIFAR100 dataset may not be enough. In this paper, we rely on the smoothness of feature representations and use the NPK $g_q$ for sample selection -- even a randomly initialized encoder can provide quite meaningful neighbouring relations therefore enabling us to train the model from scratch and also improves the sample selection stability and performance in heavy and complex noisy scenarios. For a more detailed discussion, please refer to \texttt{Supplementary C}.
%\quad Most previous works rely on the PMC $g_p$ and the memorization effect of neural networks to select clean samples, i.e., the widely-applied small-loss principle. However, it is known that such methods are not steady and fragile with complex and heavy noise. Also, these methods often require a warmup stage before sample selection which actually requires people's prior knowledge about the difficulty and noise ratio of the dataset. For example, a warmup stage of 50 epochs on the heavy noisy CIFAR10 dataset may have resulted in overfitting before sample selection, while a warmup stage of 5 epochs on the mildly noisy CIFAR100 dataset may not be enough. In this paper, we rely on the smoothness of feature representations of samples and the NPK $g_q$ for sample selection, which enables us to train the model from scratch and also improves the sample selection stability and performance in heavy and complex noisy scenarios. For more detailed discussions, please refer to \texttt{Supplementary B}.

% \paragraph{Why not NPK $g_q$ for sample relabelling?}
\textit{Why not NPK $g_q$ for sample relabelling?}
\quad Due to the existence of noisy labels, we found that it is very difficult to make a proper choice of $\theta_r$ and rely on the NPK $g_q$ for sample relabeling, especially in the early iterations. By contrast, in our scheme, PMC $g_p$ is always trained with a relatively clean subset and can perform sample relabeling more accurately. Furthermore, relabeling samples on which the classifier is confident leads to smaller gradients and smoother learning from easier samples first --- even when the newly assigned labels are wrong, the influence is smaller.

%Using PMC $g_p$ for sample relabelling is straightforward. 

%\quad Due to the existence of noisy labels, it is extremely difficult to make a proper choice of $\theta_r$ and rely on the NPK $g_q$ for sample relabeling. By contrast, in our scheme PMC $g_p$ is always trained with a relatively clean subset that can perform sample relabeling more accurately. Using PMC $g_p$ for sample relabelling is straightforward.

\subsection{Model training}
\label{3_4}
In the training stage, we use the most basic form of supervised learning, i.e., using the cross-entropy loss on the clean subset selected in the first stage --- this updates both the encoder $f$ and the PMC $g_p$. With our sample relabelling mechanism, the size of the clean subset grows progressively by including more and more relabeled closed-set noise in training. Optionally, we use a feature consistency loss that enforces consistency between the feature representations of different augmentations of the same sample --- this updates the encoder $f$ and helps to learn a strong feature space on which the selection mechanism of the first stage can rely. %\cfnote{not sure if we need this para...}

\paragraph{Supervised training using the clean subset}
%With each specific sample $(\vx, \vy^r)$ in the selected subset $(\mathcal{X}_c,\ \mathcal{Y}^r_c)$, we train the encoder $f$ and classifier $g$ with common cross-entropy loss. Meanwhile, due to possible class imbalance in the selected subset, we simply over-sample minority classes to ensure the number of samples same as majority classes. Along with the balanced sample selection in our sample selection mechanism, we investigate and report the effect of balancing strategies in \texttt{Supplementary C}.
For each sample $(\vx, \vy^r)$ in the selected subset $(\mathcal{X}_c,\ \mathcal{Y}^r_c)$, we train the encoder $f$ and PMC $g_p$ with common cross-entropy loss, that is, $L_{ce} = - {\vy^r}^T \log g_p(f(\vx))$.
% \begin{equation} \label{eq8}
%     L_{ce} = - {\vy^r}^T \log g_p(f(\vx))
% \end{equation}
Moreover, to deal with the possible class imbalance in the selected subset, we simply over-sample minority classes. In the ablations study, we report the effect of balancing -- the over-sampling and also the balanced sample selection in~\cref{eq3}.
\paragraph{Optional: feature consistency regularization using all samples} 
Although our relabeling method can progressively relabel and introduce closed-set noise samples into training, open-set samples can also improve generalization. Motivated by commonly used prediction consistency regularization methods, we propose a feature consistency loss $L_{fc}$~\citep{simsiam}. Specifically, with a projector $h_{proj}$ and predictor $h_{pred}$, we minimize the cosine distance\footnote{In \texttt{Supplementary D} we investigate the use of the L2 distance.} between two different augmented views ($\vx_1$ and $\vx_2$) of the same sample $\vx$. That is,
\begin{equation} \label{eq9}
    L_{fc} = -\frac{\vh_1^\top \vh_2}{ \| \vh_1\|_{\small 2} \| \vh_2\|_{\small 2}},
\end{equation}
where $\vh_1 \triangleq h_{pred}(h_{proj}(f(\vx_1)))$ and $\vh_2 \triangleq h_{proj}(f(\vx_2))$.%, we minimize the cosine distance between $\vh_1$ and $\vh_2$: 
%We also investigate the use of the L2 loss as the distance metric ---  details can be found in \texttt{Supplementary D}. 
In summary, the overall training objective is to minimize a weighted sum of $L_{ce}$ and $L_{fc}$, that is 
\begin{equation} \label{eq10}
    L = L_{ce}+ \lambda L_{fc}.
\end{equation}
We set $\lambda=1$. For brevity, we name our method as SSR when $\lambda=0$, and SSR+ when $\lambda \neq 0$.

\section{Experiments}
\label{4}

\subsection{Overview}
\label{4_1}
In this section, we conduct extensive experiments on two standard benchmarks with synthetic label noise, CIFAR-10 and CIFAR-100, and three real-world datasets, Clothing1M~\citep{clothing1mdataset}, WebVision~\citep{webvision}, and ANIMAL-10N~\citep{selfie}. For brevity, we define abbreviated names for the corresponding noise settings, such as "$\mathsf{sym50}$" for 50\% symmetric noise, "$\mathsf{asym40}$" for 40\% asymmetric noise and "$\mathsf{all30\_open50}$" for 30\% total noise with 50\% open-set noise.~(more dataset and implementation details can be found in \texttt{Supplementary A and B}). In~\cref{4-2}, we conduct extensive ablation experiments to show the great performance and robustness of our sample selection and relabelling mechanism w.r.t its hyperparameters with different noise types, noise ratios and dataset. In~\cref{4-3} and~\cref{4-4}, we compare our method with the state-of-the-art in synthetic noisy datasets and real-world noisy datasets. 

\subsection{Ablations study}
\label{4-2}

\begin{figure}[htbp]
\begin{center}
    \includegraphics[width=1.0\textwidth]{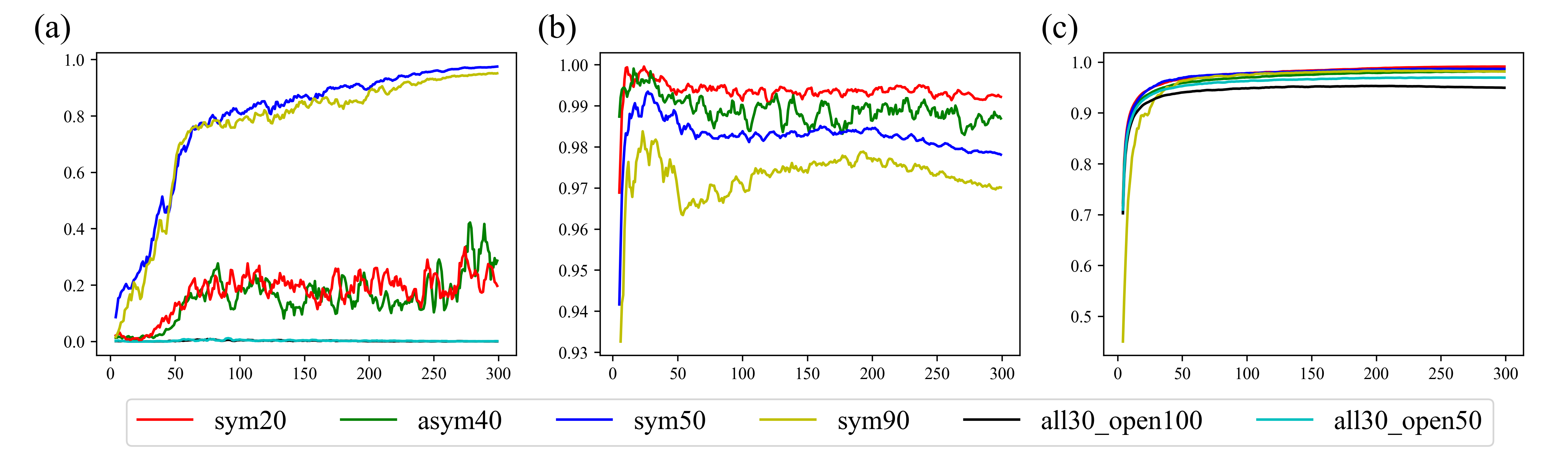}
\end{center}
\caption{Effect of our sample selection and relabelling method with various noise settings. (a) The proportion of relabeled samples; (c) The corrected clean samples ratio within the relabeled part; (c)F-score of sample selection.}
\label{fig:selectionandrelabeling}
\end{figure}

\paragraph{Quality of sample selection and relabelling} In~\cref{fig:selectionandrelabeling}, we investigate the quality of sample selection and relabeling under different noise types and ratios on the CIFAR10 noisy dataset. We set $\theta_r = 0.9$ for $\mathsf{40asym}$, $\mathsf{20sym}$, $\mathsf{all30\_open50}$ and $\mathsf{all30\_open100}$ noise, while $\theta_r = 0.8$ for $\mathsf{sym50}$ and $\mathsf{sym90}$ noise -- please refer to \texttt{Supplementary A} for more details on noise. We set $\theta_s = 1$ in all experiments.

% \begin{figure}
% \begin{center}
% \begin{subfigure}{0.3\textwidth}
%     \includegraphics[width=\textwidth]{figures/total_labels.png}
%     \caption{}
%     \label{fig:number}
% \end{subfigure}
% % \hfill
% \begin{subfigure}{0.3\textwidth}
%     \includegraphics[width=\textwidth]{figures/correct_labels.png}
%     \caption{}
%     \label{fig:correct}
% \end{subfigure}
% % \hfill
% \begin{subfigure}{0.3\textwidth}
%     \includegraphics[width=\textwidth]{figures/fscore.png}
%     \caption{}
%     \label{fig:fscore}
% \end{subfigure}
% \end{center}
% \caption{Effect of our sample selection and relabelling method with various noise settings. (a)The proportion of relabeled samples; (c)The corrected clean samples ratio within the relabeled part; (d)F-score of sample selection.}
% \label{fig:selectionandrelabeling}
% \end{figure}

To summarise, we found that the number of the relabeled samples is highly related to the value of $\theta_r$ across different noise ratios (\cref{fig:selectionandrelabeling}(a)) for closed-set noise only dataset, for e.g, a lower $\theta_r$ leads to more relabeled samples across different noise settings~($\mathsf{40asym}$, $\mathsf{20sym}$, $\mathsf{50sym}$ and $\mathsf{90sym}$). For datasets containing also open-set noise~($\mathsf{all30\_open50}$ and $\mathsf{all30\_open100}$), our relabeling mechanism is more conservative~(nearly no relabelling), thus alleviating the negative impact of open-set noise. In~\cref{fig:selectionandrelabeling}(b), we show that we obtain very high relabelling accuracy with different noise and relabelling volumes, e.g., only $19\%$ samples have correct labels originally for $90\%$ symmetric noise while >95\% samples are correctly relabelled. In~\cref{fig:selectionandrelabeling}(c), we report the F-score of our sample selection. Please note that our sample selection is more challenging compared to previous sample selection methods. While previous methods usually focus on identifying clean subsets and noisy subsets based on the original labels, our method involves the relabeling of samples, which takes the risk of introducing more errors while increasing the number of clean subsets. Even so, we see that the F-score of our sample selection works well~(> 0.95 in most cases).

\paragraph{Robustness to hyper-parameters} In this section, we conduct extensive ablation studies to show the robustness of the values of the few hyperparameters with different noise types, noise ratios, and datasets. The choice of $\theta_r$ controls the sample relabelling quality and proportion. Roughly speaking, the lower the $\theta_r$, the more samples will be relabelled. Similarly, the choice of $\theta_s$ controls the sample selection quality and proportion -- the lower the $\theta_s$, the more samples will be selected for the training stage. We set $\theta_s = 1$ for $\theta_r$ ablations and $\theta_r = 0.9$ for $\theta_s$ ablations. We also investigate the effects of $K$ --- the size of neighborhood of NPK $g_q$ during the sample selection stage, with $\theta_s = 1$ and $\theta_r = 0.9$.
\begin{figure}[htbp]
\begin{center}
\includegraphics[width=1.0\textwidth]{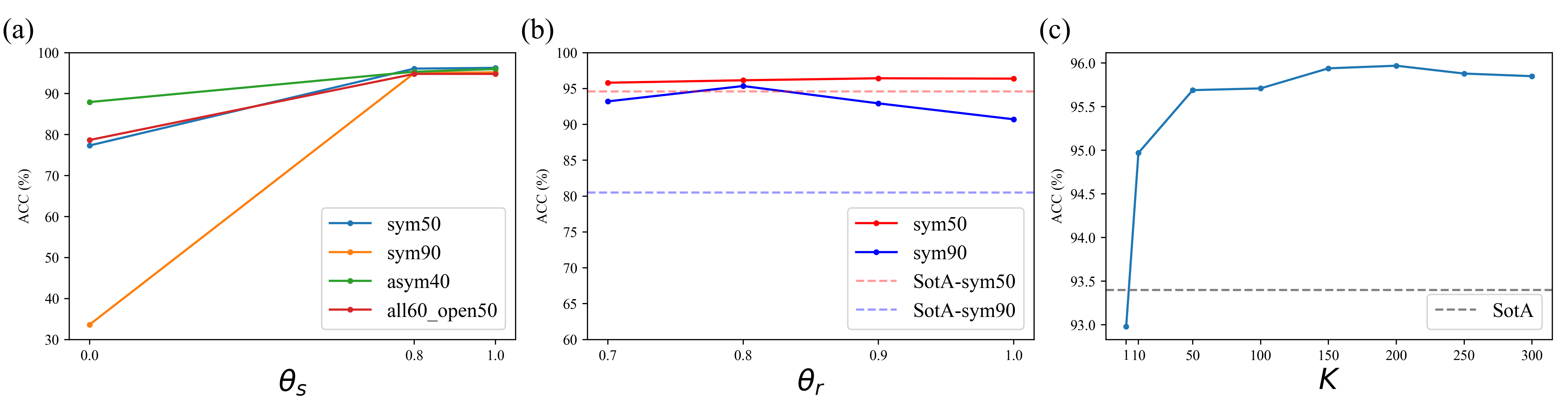}
% \begin{subfigure}{0.3\textwidth}
%     \includegraphics[width=\textwidth]{figures/s.png}
%     \caption{$\theta_s$}
%     \label{fig:thetas}
% \end{subfigure}
% % \hfill
% \begin{subfigure}{0.3\textwidth}
%     \includegraphics[width=\textwidth]{figures/thetar.png}
%     \caption{$\theta_r$}
%     \label{fig:thetar}
% \end{subfigure}
% % \hfill
% \begin{subfigure}{0.3\textwidth}
%     \includegraphics[width=\textwidth]{figures/k.png}
%     \caption{$K$}
%     \label{fig:k}
% \end{subfigure}
\end{center}
\caption{Classification accuracy with different hyper-parameters on CIFAR10 datasets. (a)$\theta_s = [0, 0.8, 1.0]$; (b)$\theta_r = [0.7,0.8,0.9,1]$; (c) $K=[1,10,50,100,150,200,250,300]$.}
\label{fig:hyperparameters}
\end{figure}
% Especially, we have four main settings:
% \begin{itemize}[itemsep=0em]
%     \item $\theta_r = 1, \theta_s = 0$ Normal supervised training
%     \item $\theta_r < 1, \theta_s = 0$ Sample relabelling only
%     \item $\theta_r = 1, \theta_s > 0$ Sample selection only
%     \item $\theta_r < 1, \theta_s > 0$ Our default setting
% \end{itemize}

In~\cref{fig:hyperparameters}(a) we report results with $\theta_s = [0, 0.8, 1.0]$. Removing sample selection ($\theta_s=0$) leads to severe degradation especially for a high noise ratio (90\% symmetric noise), while a relatively high $\theta_s$ gives consistently high performance. In~\cref{fig:hyperparameters}(b), we report the performance with different $\theta_r$ on the synthetic CIFAR10 noisy dataset. Our method achieves consistently superior performance than the state-of-the-art with different $\theta_r$. In~\cref{fig:hyperparameters}(c), we report results with different $K$ for the CIFAR10 dataset with 40\% asymmetric noise since it is more challenging and realistic. Except for extremely small $K$, our method is stable and consistently better than the state-of-the-art.

\paragraph{Effect of balancing strategies}
\label{effect}
To alleviate the possible class imbalance in the dataset, we proposed two balancing strategies, one in the sample selection~[\cref{eq3}] and one in the model training stage~[Over-sampling minority class], respectively. In~\cref{Balancing_trciks} we investigate the effect of using data balancing or not, on CIFAR10 with 40\% asymmetric noise and also on a well-known real-world imbalanced noisy dataset, Clothing1M. It can be seen that the effect is small but positive.

\begin{table}[htbp]
\begin{center}
\resizebox{0.45\textwidth}{!}{%
\begin{tabular}[t]{@{}lcc@{}}
\toprule
Method           & Clothing1M                      & 40\% asym CIFAR10               \\ \midrule
SSR               & \textbf{74.83} & \textbf{95.5} \\
w/o balancing & 74.12                           & 94.9                           \\ \bottomrule
\end{tabular}
}
\end{center}
\caption{Effect of class balancing.}
\label{Balancing_trciks}
\end{table}

\subsection{Evaluation with synthetic noisy datasets}
\label{4-3}
In this section, we compare our method to the most recent state-of-the-art methods and we show that it achieves consistent improvements in all datasets and at all noise types and ratios.

\begin{table*}[htbp]
\begin{center}
\resizebox{0.75\textwidth}{!}{
\begin{tabular}{@{}l|ccccc|cccc@{}}
\toprule
Dataset                                & \multicolumn{5}{c|}{CIFAR10}                                                                                                                                      & \multicolumn{4}{c}{CIFAR100}                                                                                                 \\ \midrule
Noise type                             & \multicolumn{4}{c|}{Symmetric}                                                                                                                    & Assymetric    & \multicolumn{4}{c}{Symmetric}                                                                                                \\ \midrule
Noise ratio                            & \multicolumn{1}{c|}{20\%}          & \multicolumn{1}{c|}{50\%}          & \multicolumn{1}{c|}{80\%}          & \multicolumn{1}{c|}{90\%}          & 40\%          & \multicolumn{1}{c|}{20\%}          & \multicolumn{1}{c|}{50\%}          & \multicolumn{1}{c|}{80\%}          & 90\%          \\ \midrule
Cross-Entropy                          & \multicolumn{1}{c|}{86.8}          & \multicolumn{1}{c|}{79.4}          & \multicolumn{1}{c|}{62.9}          & \multicolumn{1}{c|}{42.7}          & 85.0          & \multicolumn{1}{c|}{62.0}          & \multicolumn{1}{c|}{46.7}          & \multicolumn{1}{c|}{19.9}          & 10.1          \\
Co-teaching+~\citep{coteaching+}       & \multicolumn{1}{c|}{89.5}          & \multicolumn{1}{c|}{85.7}          & \multicolumn{1}{c|}{67.4}          & \multicolumn{1}{c|}{47.9}          & -             & \multicolumn{1}{c|}{65.6}          & \multicolumn{1}{c|}{51.8}          & \multicolumn{1}{c|}{27.9}          & 13.7          \\
F-correction~\citep{loss_correction}   & \multicolumn{1}{c|}{86.8}          & \multicolumn{1}{c|}{79.8}          & \multicolumn{1}{c|}{63.3}          & \multicolumn{1}{c|}{42.9}          & 87.2          & \multicolumn{1}{c|}{61.5}          & \multicolumn{1}{c|}{46.6}          & \multicolumn{1}{c|}{19.9}          & 10.2          \\
PENCIL~\citep{pencil}                  & \multicolumn{1}{c|}{92.4}          & \multicolumn{1}{c|}{89.1}          & \multicolumn{1}{c|}{77.5}          & \multicolumn{1}{c|}{58.9}          & 88.5          & \multicolumn{1}{c|}{69.4}          & \multicolumn{1}{c|}{57.5}          & \multicolumn{1}{c|}{31.1}          & 15.3          \\
LossModelling~\citep{lossmodellingbmm} & \multicolumn{1}{c|}{94.0}          & \multicolumn{1}{c|}{92.0}          & \multicolumn{1}{c|}{86.8}          & \multicolumn{1}{c|}{69.1}          & 87.4          & \multicolumn{1}{c|}{73.9}          & \multicolumn{1}{c|}{66.1}          & \multicolumn{1}{c|}{48.2}          & 24.3          \\
DivideMix*~\citep{dividemix}            & \multicolumn{1}{c|}{96.1}          & \multicolumn{1}{c|}{94.6}          & \multicolumn{1}{c|}{93.2}          & \multicolumn{1}{c|}{76.0}          & 93.4          & \multicolumn{1}{c|}{77.3}          & \multicolumn{1}{c|}{74.6}          & \multicolumn{1}{c|}{60.2}          & 31.5          \\
ELR+*~\citep{elr}                        & \multicolumn{1}{c|}{95.8}          & \multicolumn{1}{c|}{94.8}          & \multicolumn{1}{c|}{93.3}          & \multicolumn{1}{c|}{78.7}          & 93.0          & \multicolumn{1}{c|}{77.6}          & \multicolumn{1}{c|}{73.6}          & \multicolumn{1}{c|}{60.8}          & 33.4          \\
RRL~\citep{rrl}                        & \multicolumn{1}{c|}{95.8}          & \multicolumn{1}{c|}{94.3}          & \multicolumn{1}{c|}{92.4}          & \multicolumn{1}{c|}{75.0}          & 91.9          & \multicolumn{1}{c|}{79.1}          & \multicolumn{1}{c|}{74.8}          & \multicolumn{1}{c|}{57.7}          & 29.3          \\
NGC~\citep{ngc}                        & \multicolumn{1}{c|}{95.9}          & \multicolumn{1}{c|}{94.5}          & \multicolumn{1}{c|}{91.6}          & \multicolumn{1}{c|}{80.5}          & 90.6          & \multicolumn{1}{c|}{79.3}          & \multicolumn{1}{c|}{75.9}          & \multicolumn{1}{c|}{62.7}          & 29.8          \\
AugDesc*~\citep{augdesc}                & \multicolumn{1}{c|}{96.3}          & \multicolumn{1}{c|}{95.4}          & \multicolumn{1}{c|}{93.8}          & \multicolumn{1}{c|}{91.9}          & 94.6          & \multicolumn{1}{c|}{79.5}          & \multicolumn{1}{c|}{77.2}          & \multicolumn{1}{c|}{66.4}          & 41.2          \\
C2D*~\citep{c2d}                        & \multicolumn{1}{c|}{96.4}          & \multicolumn{1}{c|}{95.3}          & \multicolumn{1}{c|}{94.4}          & \multicolumn{1}{c|}{93.6}          & 93.5          & \multicolumn{1}{c|}{78.7}          & \multicolumn{1}{c|}{76.4}          & \multicolumn{1}{c|}{67.8}          & 58.7          \\ \midrule
SSR(ours)                                    & \multicolumn{1}{c|}{96.3}          & \multicolumn{1}{c|}{95.7}          & \multicolumn{1}{c|}{95.2}          & \multicolumn{1}{c|}{94.6}          & 95.1          & \multicolumn{1}{c|}{79.0}              & \multicolumn{1}{c|}{75.9}              & \multicolumn{1}{c|}{69.5}              &  61.8             \\
SSR+(ours)                                   & \multicolumn{1}{c|}{\textbf{96.7}} & \multicolumn{1}{c|}{\textbf{96.1}} & \multicolumn{1}{c|}{\textbf{95.6}} & \multicolumn{1}{c|}{\textbf{95.2}} & \textbf{95.5} & \multicolumn{1}{c|}{\textbf{79.7}} & \multicolumn{1}{c|}{\textbf{77.2}} & \multicolumn{1}{c|}{\textbf{71.9}} & \textbf{66.6} \\ \bottomrule
\end{tabular}
}
\end{center}
\caption{Evaluation on CIFAR-10 and CIFAR-100 with closed-set noise. Methods marked with an asterisk employ semi-supervised learning, model co-training or model pre-training.}
\label{close}
\end{table*}

\paragraph{Evaluation with controlled closed-set noise}
In this section, we compare SSR/SSR+ to the most competitive recent works. Table~\ref{close} shows results on CIFAR10 and CIFAR100 --- we note again for SSR/SSR+ this is without the use of model cotraining or pre-training. It is clear that our method far outperforms them~(e.g. 66.6\% accuracy on CIFAR100 with 90\% symmetric noise), not only in the case of symmetric noise but also in the more realistic asymmetric synthetic noise settings.

\begin{table*}[t]
\begin{center}
\resizebox{0.9\textwidth}{!}{%
% \begin{tabular}{lccccc|cccc}
\begin{tabular}{@{}cccc|cccc|c@{}}
\toprule
CE    & F-correction \citep{loss_correction} & ELR \citep{elr} & RRL \citep{rrl} & C2D* \citep{c2d} & DivideMix* \citep{dividemix} & ELR+* \citep{elr} & AugDesc* \citep{augdesc} & SSR+(ours)     \\ \midrule
69.21 & 69.84                                & 72.87           & 74.30           & 74.84            & 74.76                        & 74.81             & \textbf{75.11}           & \textbf{74.83} \\ \bottomrule
\end{tabular}
}
\end{center}
\caption{Testing accuracy~(\%) on Clothing1M (methods with * utilized model cotraining).}
\label{clothing1m}
\end{table*}

\paragraph{Evaluation with combined open-set noise and closed-set noise}
Table~\ref{openset} shows the performance of our method in a more complex combined noise scenario. Previous methods that are specially designed for open-set noise degrade rapidly when the open-set noise ratio is decreased from 1 to 0.5~\citep{rog,iterativeopenset}. Also, the performance of the method without considering open-set noise like DivideMix~\citep{dividemix} decreases when the open-set noise ratio is increased. EDM~\citep{evidentialmix} modifies the method of DivideMix to deal with combined noise, however, reports results that are considerably lower than ours. 

\begin{table}[htbp]
\begin{minipage}{0.5\linewidth}
\begin{center}
\resizebox{0.9\textwidth}{!}{%
\begin{tabular}{@{}l|c|cc|cc@{}}
\toprule
\multirow{2}{*}{Method}    & Noise ratio & \multicolumn{2}{c|}{0.3}                           & \multicolumn{2}{c}{0.6}                            \\ \cmidrule(l){2-6} 
                           & Open ratio  & \multicolumn{1}{c|}{0.5}           & 1             & \multicolumn{1}{c|}{0.5}           & 1             \\ \midrule
\multirow{2}{*}{ILON~\citep{iterativeopenset}}      & Best        & \multicolumn{1}{c|}{87.4}          & 90.4          & \multicolumn{1}{c|}{80.5}          & 83.4          \\
                           & Last        & \multicolumn{1}{c|}{80.0}          & 87.4          & \multicolumn{1}{c|}{55.2}          & 78.0          \\ \midrule
\multirow{2}{*}{RoG~\citep{rog}}       & Best        & \multicolumn{1}{c|}{89.8}          & 91.4          & \multicolumn{1}{c|}{84.1}          & 88.2          \\
                           & Last        & \multicolumn{1}{c|}{85.9}          & 89.8          & \multicolumn{1}{c|}{66.3}          & 82.1          \\ \midrule
\multirow{2}{*}{DivideMix~\citep{dividemix}} & Best        & \multicolumn{1}{c|}{91.5}          & 89.3          & \multicolumn{1}{c|}{91.8}          & 89.0          \\
                           & Last        & \multicolumn{1}{c|}{90.9}          & 88.7          & \multicolumn{1}{c|}{91.5}          & 88.7          \\ \midrule
\multirow{2}{*}{EDM~\citep{evidentialmix}}       & Best        & \multicolumn{1}{c|}{94.5}          & 92.9          & \multicolumn{1}{c|}{93.4}          & 90.6          \\
                           & Last        & \multicolumn{1}{c|}{94.0}          & 91.9          & \multicolumn{1}{c|}{92.8}          & 89.4          \\ \midrule
\multirow{2}{*}{SSR(ours)}       & Best        & \multicolumn{1}{c|}{96.0}              &  95.7             & \multicolumn{1}{c|}{93.8}              &  93.1             \\
                           & Last        & \multicolumn{1}{c|}{95.9}              & 95.6              & \multicolumn{1}{c|}{93.7}              &  93.1          \\ \midrule
\multirow{2}{*}{SSR+(ours)}      & Best        & \multicolumn{1}{c|}{\textbf{96.3}} & \textbf{96.1} & \multicolumn{1}{c|}{\textbf{95.2}} & \textbf{94.0} \\
                           & Last        & \multicolumn{1}{c|}{\textbf{96.2}} & \textbf{96.0} & \multicolumn{1}{c|}{\textbf{95.2}} & \textbf{93.9} \\ \bottomrule
\end{tabular}}
\end{center}
\caption{Evaluation on CIFAR10 with combined noise.}
\label{openset}

\end{minipage}
\hfill
\begin{minipage}{0.5\linewidth}
\begin{center}
\resizebox{0.9\textwidth}{!}{%
\begin{tabular}{@{}l|cc|cc@{}}
\toprule
\multirow{2}{*}{Methods}    & \multicolumn{2}{c}{WebVision}   & \multicolumn{2}{c}{ILSVRC2012}  \\ \cmidrule(l){2-5} 
                            & Top1           & Top5           & Top1           & Top5           \\ \midrule
Co-teaching~\citep{coteaching}                 & 63.58          & 85.20          & 61.48          & 84.70          \\
DivideMix~\citep{dividemix} & 77.32          & 91.64          & 75.20          & 90.84          \\
ELR+~\citep{elr}            & 77.78          & 91.68          & 70.29          & 89.76          \\
NGC~\citep{ngc}             & 79.16          & 91.84          & 74.44          & 91.04          \\
% FaMUS~\citep{famus}         & 79.40          & 92.80          & 77.00          & 92.76          \\
LongReMix~\citep{cordeiro2021longremix}                   & 78.92          & 92.32          & -              & -              \\
RRL~\citep{rrl}             & 76.3           & 91.5           & 73.3           & 91.2           \\
SSR+(ours)                  & \textbf{80.92} & \textbf{92.80} & \textbf{75.76} & \textbf{91.76} \\ \bottomrule
\end{tabular}}
\end{center}
\caption{Testing accuracy~(\%) on Webvision.}
\label{webvision}

\begin{center}
\resizebox{0.9\textwidth}{!}{%
\begin{tabular}{@{}ccccc@{}}
\toprule
Cross-Entropy  & \begin{tabular}[c]{@{}c@{}}SELFIE  \\ \citep{selfie}\end{tabular} & \begin{tabular}[c]{@{}c@{}}PLC \\ \citep{PLC}\end{tabular} & \begin{tabular}[c]{@{}c@{}}NCT  \\ \citep{nestedcoteaching}\end{tabular} & SSR+(ours)              \\ \midrule
% 79.4 $\pm$ 0.1 & 81.8 $\pm$ 0.1                                                    & 83.4 $\pm$ 0.4                                             & 84.1 $\pm$ 0.1                                                           & \textbf{88.5 $\pm$ 0.1} \\ \bottomrule
79.4 & 81.8                                                   & 83.4                                            & 84.1                                                           & \textbf{88.5} \\ \bottomrule
\end{tabular}}
\end{center}
\caption{Testing accuracy on ANIMAL-10N. }
\label{animal10n}
\end{minipage}
\end{table}

\subsection{Evaluation with real-world noisy datasets}
\label{4-4}
% Most works often focus on the noisy dataset with only close-set noise. However, it is widely acknowledged that the proportion of open-set noise is also high in real-world datasets. Some works try to augment the current sample selection framework with an extra open-set sample detector, such as in EDM. These methods usually consist of several complicated modules. 
Finally, in~\cref{clothing1m},~\cref{webvision} and~\cref{animal10n} we show results on the Clothing1M, WebVision and ANIMAL-10N datasets, respectively. To summarize, our method achieves better or competitive performance in relation to the current state-of-the-art in both large-scale web-crawled datasets and small-scale human-annotated noisy datasets.

\section{Conclusions}
\label{5}

In this paper we propose an efficient \textit{\textbf{S}ample \textbf{S}election and \textbf{R}elabelling}~(SSR) framework for \textit{\textbf{L}earning with \textbf{U}nknown \textbf{L}abel \textbf{N}oise}~(\textbf{LULN}). Unlike previous methods that try to integrate many different mechanisms and regularizations, we strive for a concise, simple and robust method. The proposed method does not utilize complicated mechanisms such as semi-supervised learning, model co-training and model pre-training, and is shown with extensive experiments and ablation studies to be robust to the values of its few hyper-parameters, and to consistently and by large surpass the state-of-the-art in various datasets. 

\vspace{5pt}
{\setlength{\parindent}{0cm}
\textbf{Acknowledgments:} This work was supported by the EU H2020 AI4Media No. 951911 project.
}

\appendix
\section*{Supplementary A: Dataset details}
\label{sup:a}
\paragraph{Synthetic noisy dataset}
CIFAR10 and CIFAR100 both consist of 50K images. Following the standard practice, for CIFAR10 and CIFAR100, we evaluate our method with two types of artificial noise: \textit{symmetric noise} by randomly replacing labels of all samples using a uniform distribution; and \textit{asymmetric noise} by randomly exchanging labels of visually similar categories, such as Horse $\leftrightarrow$ Deer and Dog $\leftrightarrow$ Cat. For closed-set noise only dataset, we test with 20\%, 50\%, 80\% and 90\% symmetric noise and 40\% asymmetric noise following DivideMix~\citep{dividemix}. For datasets including also open-set noise, following settings in EDM~\citep{evidentialmix}, we test with 30\%, 60\% total noise ratio and 50\%, 100\% open-set noise ratio on CIFAR10 dataset. The total noise ratio denotes the total proportion of noisy samples in the population of samples while the open-set noise ratio denotes the proportion of open-set noise in the noisy samples. The closed-set noise is generated as \textit{symmetric noise} while the open-set noise is randomly sampled from CIFAR100.

\paragraph{Real-world noisy dataset}
WebVision~\citep{webvision} is a large-scale dataset of 1000 classes of images crawled from the Web. Following previous work~\citep{mentornet, dividemix, moit}, we compare baseline methods on the top 50 classes from Google images Subset of WebVision. The noise ratio is estimated to be around~20\%. 
ANIMAL-10N~\citep{selfie} is a smaller and recently proposed real-world dataset consisting of 10 classes of animals, that are manually labelled with an error rate that is estimated to be approximately~8\%. ANIMAL-10N has similar size characteristics to the CIFAR datasets, with 50000 train images and 10000 test images.
Clothing1M~\citep{clothing1mdataset} is a large-scale dataset of 14 classes of clothing images crawled from online shopping websites, consisting of 1 million noisy images. The noise ratio is estimated to be around~38.5\%. 

\section*{Supplementary B: Implementation details}
\label{sup:b}
We use a PresActResNet-18~\citep{preactresnet} as the backbone for all CIFAR10/100 experiments following previous works. Unlike previous methods that use specific warmup settings for CIFAR10/CIFAR100, we train the model from scratch with $\theta_s = 1.0$ in all experiments. We set $\theta_r=0.8$ for higher noise ratio --- $\mathsf{sym50},\mathsf{sym80}$ and $\mathsf{sym90}$ noise, $\theta_r=0.9$ for remain settings in all CIFAR experiments except in the corresponding ablation part. We train all modules with the same SGD optimizer for 300 epochs with a momentum of 0.9 and a weight decay of 5e-4. The initial learning rate is 0.02 and is controlled by a cosine annealing scheduler. The batchsize is fixed as 128. 

For WebVision, we use InceptionResNetv2 following~\citep{dividemix}. We train the network with SGD optimizer for 150 epochs with a momentum of 0.9 and a weight decay of 1e-4. The initial learning rate is 0.01 and reduced by a factor of 10 after 50 and 100 epochs. The batchsize is fixed as 32. 
For Clothing1M, we use ResNet50 following~\citep{dividemix} with ImageNet pretrained weights. We train the network with SGD optimizer for 150 epochs with a momentum of 0.9 and weight decay of 1e-3. The initial learning rate is 0.002 and reduced by a factor of 10 after 50 and 100 epochs. The batchsize is fixed as 32. 
For ANIMAL-10N, we use VGG-19~\citep{vgg19} with batch-normalization following~\citep{selfie}. We train the network with SGD optimizer for 150 epochs with a momentum of 0.9 and weight decay of 5e-4. The initial learning rate is 0.02 and reduced by a factor of 10 after 50 and 100 epochs. The batchsize is fixed as 128. For all real-world noisy datasets, we train the model from scratch with $\theta_s=1$, while $\theta_r$ is fixed as 0.95.

Following recent works, in this work, we define three augmentation strategies: original image which we denote with `none' augmentation for testing, random cropping+horizontal flipping which we denote as `weak' augmentation, and `strong' augmentation  the one that further combines the augmentation policy from \citep{autoaugment}. For $L_{ce}$ we use 'strong' augmentation with mixup interpolations~\citep{Mixup} while for $L_{fc}$, we use 'weak' augmentation for $\vx_2$ and 'strong' augmentation for $\vx_1$ in eq.(4).
For mixup interpolation, following DivideMix~\citep{dividemix}, we set $\alpha, \beta$ as 4 for beta mixture for the CIFAR10/CIFAR100 datasets, and as 0.5 for the real-world noisy dataset. 

\section*{Supplementary C: Evaluation on different choices for sample selection}
\label{sup:c}

In this section, we extensively compare the performance of different sample selection mechanisms under different noise ratios and modes. More specifically, with the PMC $g_p$ and NPK $g_q$, we compare the performance of these two classifiers in two different selection modes. Pre-defined mode means the number of noisy samples is known. That is, given a symmetric noisy dataset with noise ratio as $\tau$, we select the top $1-\tau$ percent of the samples as clean according to its loss value or prediction confidence. Automatic mode means that there is no information about the ratio of noise. In this case, we utilize the consistency measure $c$ using NPK in our method and the GMM-based loss modelling for PMC. Note that here we only considered two simple modes for sample selection and one variant in each mode respectively. There are many different variants proposed, however, we just aim to show the robustness of NPK over PMC here.  For a fair and clear comparison, we also show results of training with the whole dataset and clean subset as the bottom baseline and top baseline, respectively. Note, that in order to compare only the effect of the sample selection part, we exclude sample relabeling, strong data augmentation and optional feature consistency loss here.

\begin{table}[htbp]
\begin{center}
\resizebox{1.0\textwidth}{!}{
\begin{tabular}{@{}lllccccc@{}}
\toprule
Methods                      &                            &      & 50\% sym       & 80\% sym        & 40\% asym      & 60\% all(50\% open) & 60\% all(100\% open) \\ \midrule
\multirow{4}{*}{Automatic}   & \multirow{2}{*}{PMC}       & Last & 88.41          & 52.19          & 48.88          & 83.91               & \textbf{85.24}       \\ \cmidrule(lr){3-3}
                             &                            & Best & 88.52          & 64.03          & 79.24          & 84.08               & \textbf{85.32}       \\ \cmidrule(l){2-8} 
                             & \multirow{2}{*}{NPK(Ours)} & Last & \textbf{88.82} & \textbf{68.67} & \textbf{88.20} & \textbf{85.27}      & 84.72                \\ \cmidrule(lr){3-3}
                             &                            & Best & \textbf{88.88} & \textbf{69.57} & \textbf{88.99} & \textbf{85.30}      & 85.17                \\ \midrule
\multirow{4}{*}{Pre-defined} & \multirow{2}{*}{PMC}       & Last & 87.85          & 65.93          & \textbf{84.33} & 85.75               & 86.16                \\ \cmidrule(lr){3-3}
                             &                            & Best & 87.98          & 66.33          & \textbf{85.41} & 85.76               & 86.48                \\ \cmidrule(l){2-8} 
                             & \multirow{2}{*}{NPK}       & Last & \textbf{88.20} & \textbf{72.25} & 75.47          & \textbf{85.84}      & \textbf{86.43}       \\ \cmidrule(lr){3-3}
                             &                            & Best & \textbf{88.27} & \textbf{72.66} & 83.36          & \textbf{86.10}      & \textbf{86.53}       \\ \midrule
Whole dataset                &                            & Last & 58.93          & 27.59          & 76.68          & 59.92               & 80.99                \\ \cmidrule(lr){3-3}
                             &                            & Best & 80.86          & 61.55          & 85.41          & 79.26               & 84.18                \\ \midrule
Clean subset                 &                            & Last & 92.46          & 87.64          & 93.40          & 90.70               & 90.62                \\ \cmidrule(lr){3-3}
                             &                            & Best & 92.53          & 87.85          & 93.56          & 90.87               & 90.82                \\ \bottomrule
\end{tabular}
}
\end{center}
\caption{Results with different sample selection mechanisms.}
\label{tab:sampleselection}
\end{table}

In~\cref{tab:sampleselection}, we can see that the NPK-based selection achieved better performance compared to PMC-based selection regardless of the mode, and that our choice can significantly improve the baseline~(Whole dataset) without knowing the noise ratio. 

\section*{Supplementary D: Distance metric for feature consistency loss}
\label{sup:d}
In Section 3.3, we use the cosine distance as the distance metric for feature consistency loss. Here we also experiment with the use of the L2 distance. The results are shown in~\cref{distancemetric}, where it can be seen that there are small differences, but that the cosine similarity is in general better.

\begin{table}[htbp]
\begin{center}
\resizebox{0.9\textwidth}{!}{%
\begin{tabular}{@{}lcccc@{}}
\toprule
Training                       & 50\%~sym       & 90\%~sym        & 40\%~asym      & 60\% all~(50\% open-set) \\ \midrule
SSR+(negative cosine similarity) & \textbf{96.1} & \textbf{95.2} & 95.5           & \textbf{95.2}          \\
SSR+(L2 distance)                & 96.0          & 94.7           & \textbf{96.1} & 94.3                   \\ \bottomrule
\end{tabular}}
\end{center}
\caption{SSR+ with different distance metric for feature consistency loss}
\label{distancemetric}
\end{table}

\section*{Supplementary E: Computational cost analysis}
\label{sup:e}
In~\cref{tab:time}, we report the running time of each step of our model on the datasets that we have experimented with. It can be seen that the time of sample selection and relabelling is almost negligible compared to the time of gradient propagation.

% \newpage
\begin{table}[htbp]
\begin{center}
\resizebox{0.8\textwidth}{!}{
\begin{tabular}{@{}lcccc@{}}
\toprule
\multirow{2}{*}{Dataset(Size)} & \multirow{2}{*}{Model training} & \multicolumn{3}{c}{Sample selection and relabelling}       \\ \cmidrule(l){3-5} 
                               &                                 & Feature extraction & sample selection & sample relabelling \\ \midrule
CIFAR(50K)                     & 112s                            & 9s                 & 1.23s            & $\sim$~0s                 \\
WebVision(~65K)                & 587s                            & 109s               & 1.48s            & $\sim$~0s                \\
Clothing1M(32K)                & 575s                            & 57s                & 0.79s            & $\sim$~0s                \\ \bottomrule
\end{tabular}}
\end{center}
\caption{Computational cost analysis.}
\label{tab:time}
\end{table}

\end{document}